\RequirePackage[T1]{fontenc}
\documentclass[conference]{IEEEtran}
\IEEEoverridecommandlockouts
\usepackage{cite}
\usepackage{amsmath,amssymb}
\usepackage{graphicx}
\usepackage{booktabs}
\usepackage{url}

\usepackage{xcolor}
\graphicspath{{figures/}}
\usepackage{eso-pic}
\AddToShipoutPictureBG*{\AtPageUpperLeft{\put(0,-14){\makebox[\paperwidth]{\parbox[t]{0.86\paperwidth}{\centering\scriptsize
\copyright~2026 IEEE. Personal use of this material is permitted. Permission from IEEE must be obtained for all other uses, in any current or future media, including reprinting/republishing this material for advertising or promotional purposes, creating new collective works, for resale or redistribution to servers or lists, or reuse of any copyrighted component of this work in other works.}}}}}

\providecommand{\revmarked}{0}
\ifnum\revmarked=1 \DeclareRobustCommand{\rev}[1]{\textcolor{blue}{#1}}\else \newcommand{\rev}[1]{#1}\fi

\begin{document}

\title{Chunk Recomputation under Real Document Revisions:\\ Measuring Update Cost and Retrieval Quality of Fixed, Recursive, and Content-Defined Chunking}

\newif\ifanonymous
\anonymousfalse
\ifanonymous
\author{}
\else
\author{\IEEEauthorblockN{Yuning Bian\textsuperscript{*}}
\IEEEauthorblockA{\textit{School of Computer Science}\\
\textit{and Technology}\\
\textit{Anhui University of Technology}\\
Ma'anshan, China\\
bianyuning@gmail.com}
\and
\IEEEauthorblockN{Lele Xu}
\IEEEauthorblockA{\textit{School of Computer Science}\\
\textit{and Technology}\\
\textit{Anhui University of Technology}\\
Ma'anshan, China\\
jixu0090@gmail.com}
\and
\IEEEauthorblockN{Yifan Yuan}
\IEEEauthorblockA{\textit{School of Computer Science}\\
\textit{and Technology}\\
\textit{Anhui University of Technology}\\
Ma'anshan, China\\
yyifan355@gmail.com}
\and[\hfill\mbox{}\par\mbox{}\hfill]
\IEEEauthorblockN{Xinyu Yin}
\IEEEauthorblockA{\textit{School of Computer Science}\\
\textit{and Technology}\\
\textit{Anhui University of Technology}\\
Ma'anshan, China\\
Stressful0614@outlook.com}
\and
\IEEEauthorblockN{Chengyan Zhan}
\IEEEauthorblockA{\textit{School of Computer Science}\\
\textit{and Technology}\\
\textit{Anhui University of Technology}\\
Ma'anshan, China\\
zhan82789@gmail.com}
\thanks{\textsuperscript{*}Yuning Bian, Lele Xu, and Yifan Yuan contributed equally to this work and share first authorship.}}
\fi

\IEEEaftertitletext{\vspace{-2\baselineskip}}
\maketitle

\begin{abstract}
Chunking strategies for retrieval-augmented generation are almost always evaluated on a static snapshot, yet the documents underlying a knowledge base continue to change. Because chunks are embedded independently, a boundary that moves after an edit forces unchanged text to be re-embedded. We measure this update cost on 120 pairs of adjacent Chinese and English Wikipedia revisions for three stateless chunkers (\rev{non-overlapping} fixed-size, recursive paragraph-first, and content-defined chunking) under a strict byte-identical reuse rule and a frozen dense encoder. On the 116-article common-success pool, with frozen, length-calibrated parameters, fixed-size chunking recomputes 57--68\% of its full new-version embedding-input workload, recursive chunking 38--56\%, and content-defined chunking 43--54\% \rev{(it failed on 4 of 62 Chinese articles)}, while \rev{at these parameters} mean evidence recall under a fixed 2048-token retrieval budget differs by less than two percentage points within either language. A \rev{cost-only} parameter grid shows that the ranking of the two content-aware chunkers \rev{changes} with chunk-length settings, whereas \rev{non-overlapping} fixed-size chunking has the highest recomputation cost in every length-matched condition. Controlled edits and a verbatim-presence diagnostic support boundary drift as one contributor to recomputation of text already present in the previous version. Update cost is therefore a substantial, parameter-sensitive dimension that static retrieval evaluations do not expose and that should be reported alongside retrieval quality.
\end{abstract}
\begin{IEEEkeywords}
retrieval-augmented generation, chunking, document revision, embedding reuse, incremental indexing
\end{IEEEkeywords}

\section{Introduction}
A retrieval-augmented generation (RAG) pipeline splits each document into chunks, embeds every chunk independently, and retrieves chunks by vector similarity. Chunking strategies are usually tuned and compared on a fixed corpus~\cite{qu-etal-2025-semantic,wang-etal-2025-document,shaukat-etal-2026-chunking-sensitivity,caspari-etal-2026-complex-chunking}. Knowledge bases, however, are revised continually, and a revision reaches the index through the chunker: when a boundary moves, the chunks on either side of it become new embedding inputs even though their words remain unchanged. In our benchmark, an English article had an atom-level edit fraction of only 0.13\%, while fixed-size windows re-embedded 49\% and paragraph-first chunking re-embedded 8\%, illustrating substantial recomputation relative to the edit share.

How large this effect is on real revisions, whether it differs between chunking paradigms, and whether lower recomputation is accompanied by evidence loss during retrieval remain underexplored for natural edits with a dense encoder. Content-defined chunking (CDC) was designed to address an analogous boundary-stability problem in file systems~\cite{muthitacharoen-etal-2001-lbfs,xia-etal-2016-fastcdc}, and text adaptations such as Steadlith~\cite{steadlith-software} and KARA~\cite{kara-software} report embedding reuse, but we did not identify a directly comparable evaluation that includes standard paragraph-first chunking on real revisions under fixed retrieval conditions.

We ask three questions. \textbf{RQ1:} How much of a revised document's embedding input must be recomputed under fixed-size, recursive, and content-defined chunking? \textbf{RQ2:} What drives the difference: edit size, edit position, or boundary rules? \textbf{RQ3:} Does a chunker that recomputes less retrieve less evidence under the same encoder and token budget?

Our contributions are (i) a measurement procedure for update cost based on byte-identical reuse of complete embedding inputs, validated by hand-computed contracts and a no-edit control; (ii) a bilingual benchmark of 120 adjacent Wikipedia revision pairs stratified by edit size, with 240 questions whose version-specific evidence, when present, is anchored to character offsets; and (iii) an empirical comparison showing that, at frozen parameters, \rev{non-overlapping} fixed-size chunking differs from the content-aware methods by double-digit percentage points in mean update cost while observed mean retrieval recall differs by only a few percentage points, that the cost ranking of the two content-aware chunkers is parameter-dependent, and that the diagnostics support boundary drift as an explanation for recomputation of unchanged text.

\section{Related Work}
\textbf{Chunking for retrieval.}
Recent evaluations compare chunking strategies on static corpora. Qu et al.~\cite{qu-etal-2025-semantic} ask whether semantic chunking justifies its cost; Wang et al.~\cite{wang-etal-2025-document} show that segmentation choices change downstream answer quality; Shaukat et al.~\cite{shaukat-etal-2026-chunking-sensitivity} evaluate 36 methods across domains and embedding models; Caspari et al.~\cite{caspari-etal-2026-complex-chunking} weigh retrieval gains against indexing and query cost as the corpus grows. These studies primarily evaluate static or single-index settings. We adopt a controlled setup with one frozen embedding model and one token budget, and ask what happens when the same document is indexed again after a real revision.

\textbf{Reuse across versions.}
Content-defined chunking was introduced for file synchronisation and deduplication~\cite{muthitacharoen-etal-2001-lbfs,xia-etal-2016-fastcdc}: boundaries chosen from local content survive insertions elsewhere in the stream. Steadlith~\cite{steadlith-software} applies word-level Rabin fingerprints to text and reports reuse on synthetic edits with hashed representations; we run its pinned implementation on real revisions with a dense encoder and count exact embedding-input tokens rather than chunks. KARA~\cite{kara-software} extends this idea to text pipelines: it retains the previous chunk layout, builds a candidate-boundary graph, and chooses a shortest path that maximises reuse of existing chunks. Its objective counts chunks and padding penalties rather than the exact embedding-input tokens we measure, and it requires the previous layout as state, whereas the chunkers we compare are stateless functions of the current version.

\textbf{Index maintenance and revision corpora.}
Dynamic vector indexes such as FreshDiskANN~\cite{singh-etal-2021-freshdiskann} and SPFresh~\cite{xu-etal-2023-spfresh} update the index once vectors have changed; our question concerns the preceding step: which vectors need to change. Revision corpora such as arXivEdits~\cite{jiang-etal-2022-arxivedits}, IteraTeR~\cite{du-etal-2022-understanding-iterative}, and ARIES~\cite{darcy-etal-2024-aries} differ in source and alignment granularity. Our measure requires the exact bytes of both versions, and under a byte-identical reuse rule extraction noise is indistinguishable from an edit; in a pilot involving two versions of one arXiv paper, we found that citation handling alone produced spurious differences in unedited passages. Wikipedia revisions provide the exact text of both versions natively, so they form the retrieval benchmark; Sec.~\ref{sec:results} adds a cost-only check on arXivEdits while retaining its extraction noise as a limitation.

\section{Measuring Update Cost}
\textbf{Setting.}
A document $D$ has an old version $D_{\mathrm{old}}$ and a new version $D_{\mathrm{new}}$. A chunker $c$ partitions a version's text into an ordered sequence of chunks; every chunk is embedded independently with a frozen model, so the embedding input of a chunk is its exact text plus the model's special tokens. All three chunkers studied here are stateless: they see only the current version and fixed parameters, never the previous layout, alignment labels, or questions.

\textbf{Reuse.}
An embedding is reused only if its complete input is byte-identical to one produced for $D_{\mathrm{old}}$ under the same model configuration. This strict exact-match rule requires no alignment heuristics: two chunks that differ by a single character are two distinct inputs. The cache holds the previous version of the same document only; we do not count matches across documents.

\textbf{Recompute fraction.}
Let $T(x)$ be the number of input tokens of chunk $x$ under the frozen tokenizer, including special tokens. For chunker $c$, the recompute fraction of a revision is
\begin{equation}
  R_c = \frac{\sum_{x \in M_c} T(x)}{\sum_{x \in U_c(D_{\mathrm{new}})} T(x)},
\end{equation}
where $U_c(D_{\mathrm{new}})$ is the set of distinct complete new-version embedding inputs and $M_c \subseteq U_c(D_{\mathrm{new}})$ those absent from the previous-version cache. Both sums count each distinct input once; identical repeated chunk occurrences share one embedding computation. $R_c = 0$ means all distinct inputs are cached, whereas $R_c = 1$ means all are cache misses. Each method is normalised by its own full-new-version input workload under this deduplication policy. $R_c$ counts model input, not wall-clock time or price.

\textbf{No-edit control.}
Chunking $D_{\mathrm{old}}$ twice must yield $R_c = 0$; we run this check for every document and method and treat any non-zero result as a pipeline failure rather than as a measurement.

\textbf{Edit size.}
Formal token-level edit ratios require an audited character alignment that we do not yet have (Sec.~\ref{sec:limits}). As a proxy, we report the atom-level edit fraction: the text is split into atoms (words for Latin script, single characters for CJK, punctuation), and Python's \texttt{difflib.SequenceMatcher} is applied to the atom sequences with \texttt{autojunk=False}. The matcher recursively anchors on longest contiguous matching blocks; it does not compute an optimal longest common subsequence. From the resulting matching blocks, the fraction is $(\text{removed} + \text{added})/(|A_{\mathrm{old}}| + |A_{\mathrm{new}}|)$. Moves count as a deletion plus an insertion, so the proxy overstates edits that only relocate text. Size bands (small $\le 5\%$, medium $\le 20\%$, large $\le 50\%$, extreme $> 50\%$) are defined on this proxy before any method is run.

\textbf{Retrieval quality.}
For each question and version, we retrieve chunks from a fixed multi-document pool with the same frozen encoder, rank by exact dot product, and pack whole chunks in rank order into a 2048-input-token budget. Evidence recall is 1 if the packed chunks cover at least one sufficient evidence set specified by the label and 0 otherwise; versions without an answer are excluded rather than scored. Annotation provenance and quality checks are described in Sec.~\ref{sec:setup} and Sec.~\ref{sec:limits}.

\textbf{Validation.}
The reuse and recomputation calculations pass 20 hand-computed contracts covering insertions, deletions, replacements, moves, repeated text, and empty documents, and 13 tokenizer checks against the frozen model.

\section{Benchmark and Setup}\label{sec:setup}
\textbf{Documents.}
We use adjacent revision pairs of Chinese and English Wikipedia articles, collected in three rounds under the same frozen rules: both versions contain 1,024--20,000 tokenizer units of body text, and each pair consists of adjacent revisions in the article history. For the frozen runs reported here, reverted, truncated, off-topic, or structurally broken candidates were excluded before the reported chunking runs. The first two rounds included raw-wikitext inspection; the larger third round used automated filters followed by two separate protocol-guided model checks. Cohort decisions for the reported runs were blind to method output; failures and exclusions remain recorded. Size bands were fixed in the released protocol before experimentation: small ($\le 5\%$), medium ($\le 20\%$), large ($\le 50\%$), and extreme. In total, there are 149 pairs: 6 disjoint calibration articles, 120 primary articles, and 23 articles with unresolved source reservations retained only as a diagnostic cohort. The primary cohort has 62 Chinese articles (24 small, 22 medium, 14 large, 2 extreme) and 58 English articles (28, 12, 13, and 5); all three chunkers succeeded on 116, which form the common-success pool for paired comparisons.

\textbf{Questions and evidence.}
Each primary article has two questions (240 total, 480 version labels). A label records whether a version answers the question and, if so, one or more sufficient sets of character spans; a set may span several passages and alternatives are allowed. Human annotators assigned all 480 version-level labels based on the frozen source texts. The labels were not independently adjudicated by multiple human annotators. The questions include evidence that changes, appears, disappears, or moves between versions.

\textbf{Chunkers.}
We compare three stateless chunkers, each using only the current version and fixed parameters\rev{, on the same canonical text, in which each paragraph, heading, and list item is one whitespace-normalised block and blocks are separated by a blank line. \emph{Fixed-size} chunking cuts greedy, non-overlapping windows of at most 304 tokens (special tokens excluded) at token start offsets. \emph{Recursive} chunking keeps a span within its 400-token target and otherwise splits it, in order, at blank lines, line breaks, and Latin or full-width sentence terminators followed by whitespace, falling back to fixed-size windows; the units are then packed greedily while they fit the target. The rules are language-independent, but Chinese terminators are rarely followed by whitespace, so oversized Chinese blocks usually fall back to windows; such blocks are rare, and in a fixed 24-article development subset 555 of 566 recursive chunks end at a block boundary. \emph{CDC} uses} the pinned Steadlith implementation~\cite{steadlith-software}, whose word-level Rabin fingerprints select cut points that we project back onto the original text so that every chunker partitions the same string losslessly. \rev{No} chunk may exceed 512 content tokens. Targets were chosen on six calibration articles so that the document-balanced median chunk length was within 15\% across methods: fixed 304, recursive 400, CDC internal minimum/maximum 176/352. Parameters were then frozen. On the primary cohort the realised \rev{new-version} median ratio is 1.17 in both languages, \rev{mainly because recursive chunks are the longest, or tied for longest, in 101 of 116 articles.}

\textbf{Retrieval.}
Chunks and queries are embedded with a frozen BGE-M3 checkpoint~\cite{chen-etal-2024-m3} (CLS pooling, unit-normalised, exact dot product). For each question and version, the pool contains every article in the batch in its old version except the queried article, which appears in the queried version; the pool is fixed across chunkers. Ranked chunks are packed whole, skipping chunks that do not fit, considering at most 100 candidates. Each label is scored once per chunker in otherwise identical pools, so misses count per label--chunker pair.

\textbf{Statistics.}
On the 116-article common-success pool, 464 potential version labels yield 330 scorable labels after excluding unanswerable versions (153 Chinese and 177 English). Recompute fraction and evidence recall are averaged first within each article, then across articles in each language. Paired differences between chunkers are bootstrapped over articles (2,000 resamples, 95\% percentile interval). Intervals describe uncertainty in this development sample; we do not test hypotheses or claim non-inferiority.

\section{Results}\label{sec:results}
\begin{table}[t]
\centering
\caption{Mean recompute fraction and evidence recall@2048 on the primary common-success cohort. Frozen parameters (\rev{non-overlapping} fixed 304, recursive 400, CDC 176--352 tokens). \rev{Succ.: successful/all primary articles.} CDC failed on 4 Chinese articles; those runs are retained as NA.}
\label{tab:main}
\footnotesize
\setlength{\tabcolsep}{3.5pt}
\begin{tabular}{llrrrrr}
\toprule
Lang. & Chunker & \rev{Succ.} & $n$ & Recomp.\ (\%) & Recall (\%) & Med.\ len. \\
\midrule
Chinese & Fixed & \rev{62/62} & 58 & 68.4 & 97.4 & 304 \\
 & Recursive & \rev{62/62} & 58 & 55.7 & 98.3 & 356 \\
 & CDC & \rev{58/62} & 58 & 54.0 & 98.3 & 314 \\
\addlinespace
English & Fixed & \rev{58/58} & 58 & 57.4 & 96.7 & 304 \\
 & Recursive & \rev{58/58} & 58 & 37.9 & 96.8 & 342 \\
 & CDC & \rev{58/58} & 58 & 42.8 & 95.0 & 291 \\
\bottomrule
\end{tabular}
\end{table}

\begin{table*}[t]
\centering
\caption{Paired per-document differences at the frozen parameters (mean and 95\% bootstrap interval over documents, percentage points) and the range of the same recompute difference across length-matched parameter settings in \rev{the full-cohort cost grid of Sec.~\ref{sec:results} (15 Chinese, 14 English settings;} a spread over settings, not an interval).}
\label{tab:paired}
\small
\begin{tabular}{llrrr}
\toprule
Lang. & Contrast & $\Delta$Recompute [95\% CI] & $\Delta$Recall [95\% CI] & Grid range of $\Delta$Recompute \\
\midrule
Chinese & Recursive minus fixed & -12.7 [-19.7, -6.2] & +0.9 [-1.7, +3.4] & \rev{-16.4 to -5.9} \\
 & CDC minus fixed & -14.4 [-21.5, -7.9] & +0.9 [+0.0, +2.6] & \rev{-16.6 to -6.6} \\
 & CDC minus recursive & -1.7 [-6.8, +3.7] & +0.0 [-2.6, +2.6] & \rev{-0.7 to +5.5} \\
\addlinespace
English & Recursive minus fixed & -19.6 [-26.2, -13.2] & +0.1 [-2.3, +2.7] & \rev{-20.8 to -12.5} \\
 & CDC minus fixed & -14.7 [-20.5, -9.2] & -1.7 [-4.0, +0.0] & \rev{-16.4 to -13.7} \\
 & CDC minus recursive & +4.9 [+0.9, +9.4] & -1.9 [-4.5, +0.0] & \rev{-2.4 to +6.2} \\
\bottomrule
\end{tabular}
\end{table*}

\begin{table}[t]
\centering
\caption{Mean recompute fraction (\%) by edit-size band on the common-success pool, frozen parameters. Bands are assigned on the atom-level edit fraction before chunking (small $\le 5\%$, medium $\le 20\%$, large $\le 50\%$, extreme above). With 2 to 28 articles per band the rows are descriptive only.}
\label{tab:bands}
\footnotesize
\setlength{\tabcolsep}{4pt}
\begin{tabular}{llrrrr}
\toprule
Lang. & Band & $n$ & Fixed & Recursive & CDC \\
\midrule
Chinese & small & 23 & 62.4 & 45.3 & 36.1 \\
Chinese & medium & 21 & 71.1 & 61.3 & 62.3 \\
Chinese & large & 12 & 71.2 & 59.7 & 67.7 \\
Chinese & extreme & 2 & 91.8 & 93.1 & 92.0 \\
English & small & 28 & 46.3 & 24.7 & 27.2 \\
English & medium & 12 & 62.0 & 39.1 & 48.5 \\
English & large & 13 & 72.6 & 56.6 & 63.1 \\
English & extreme & 5 & 69.5 & 60.1 & 63.9 \\
\bottomrule
\end{tabular}
\end{table}

\begin{figure}[t]
  \centering
  \includegraphics[width=\columnwidth]{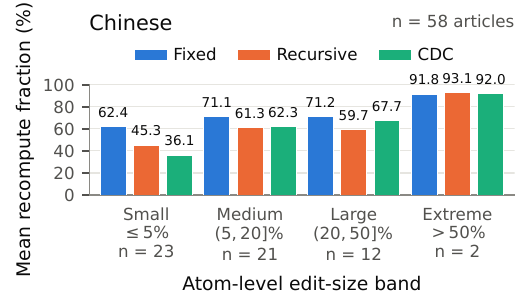}
  \caption{\rev{Mean recompute fraction by edit-size band, 58 Chinese common-success primary articles, frozen parameters (Table~\ref{tab:bands}); descriptive only.}}
  \label{fig:edit-zh}
\end{figure}
\begin{figure}[t]
  \centering
  \includegraphics[width=\columnwidth]{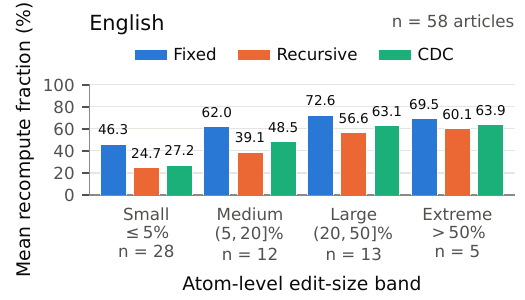}
  \caption{\rev{Mean recompute fraction by edit-size band, 58 English common-success primary articles, frozen parameters (Table~\ref{tab:bands}); descriptive only.}}
  \label{fig:edit-en}
\end{figure}

\textbf{Update cost (RQ1).}
Table~\ref{tab:main} reports means for the 116 articles processed by all three chunkers; Table~\ref{tab:paired} gives paired differences and bootstrap intervals. With frozen parameters, \rev{non-overlapping} fixed-size windows re-embed 68\% of the new input in Chinese and 57\% in English; recursive chunking 56\% and 38\%; CDC 54\% and 43\%. Recursive-minus-fixed is $-12.7$ percentage points in Chinese (95\% interval $[-19.7, -6.2]$) and $-19.6$ in English ($[-26.2, -13.2]$); CDC-minus-fixed is $-14.4$ ($[-21.5, -7.9]$) and $-14.7$ ($[-20.5, -9.2]$). All successful primary article--method runs passed the no-edit control. On the full Chinese primary cohort, fixed-size and recursive chunking average 67.1\% and 55.2\%; four CDC failures remain NA.

The differences are largest where edits are small. Among 28 small English edits, fixed-size chunking re-embeds 46\% on average, recursive chunking 25\%, and CDC 27\%; among 23 small Chinese edits the corresponding means are 62\%, 45\% and 36\%. For extreme edits the methods are closer (70\%, 60\%, 64\% in English; about 92\% in Chinese). The bands remain unbalanced (2--28 articles), so Table~\ref{tab:bands} is descriptive rather than a stratum-level estimate.

\rev{\textbf{CDC failure sensitivity.}
CDC succeeded on 116 of 120 primary articles (58/62 Chinese) and on all 23 diagnostic articles; the baselines never failed (Table~\ref{tab:main}). Imputing the four failures by full re-embedding, the overall or size-band mean of successful Chinese CDC runs, or the article's fixed-size, recursive, or larger baseline value keeps Chinese CDC-minus-fixed between $-13.5$ and $-10.1$ points with every interval below zero (full re-embedding: $[-17.9, -2.7]$), and CDC-minus-recursive between $-1.6$ and $+1.9$ with every interval including zero. The failed articles were cheaper than average for the baselines (fixed-size 48.3\% vs.\ 68.4\%), and both baselines retrieved all labelled evidence in them.}

\textbf{The content-aware ranking is parameter-dependent.}
With frozen parameters, CDC-minus-recursive is $-1.7$ percentage points in Chinese (interval $[-6.8, +3.7]$) but $+4.9$ in English ($[+0.9, +9.4]$). \rev{As the ratio of 1.17 exceeds the 1.15 criterion, we repeated the cost measurement without retuning for all primary articles at six targets per chunker (256--496 tokens; for CDC, the maximum length), keeping the 47 Chinese articles on which CDC never failed. In the 14 English and 15 Chinese length-matched settings (median-length ratio at most 1.15 in both versions),} CDC-minus-recursive spans \rev{$-2.4$ to $+6.2$ points} in English and \rev{$-0.7$ to $+5.5$ in Chinese, changing} sign in both languages\rev{; no interval lies below zero and 13 of 29 lie above it}. For example, on \rev{all 58} English articles, raising the recursive target from 400 to 448 increases recomputation from \rev{37.9\%} to \rev{46.2\%} while CDC at 448 stays at \rev{43.8\%}. Recursive packing can regroup downstream paragraphs after a length-changing edit, and a larger target allows that regrouping to propagate farther. Fixed-size chunking has the highest recomputation cost in every length-matched setting, by \rev{13.7--16.4 (English) and 6.6--16.6 (Chinese) points} against CDC\rev{; 57 of the 58 per-setting intervals of its two contrasts lie below zero, and all do under stricter gates of 1.10 and 1.05}. Thus, \rev{non-overlapping} fixed-size chunking has the highest recomputation cost across the tested grid, while the content-aware ranking is parameter- and language-dependent.

\textbf{External check on arXivEdits.}
Of 75 development papers, 63 met the frozen rules; all three chunkers succeeded on a seeded sample of 30\rev{, with mean recompute fractions of 94.5\% (fixed-size), 81.4\% (recursive), and 80.2\% (CDC)}. Recursive-minus-fixed was $-13.1$ percentage points (95\% paired interval $[-21.3, -5.9]$), CDC-minus-fixed $-14.3$ $[-22.1, -7.6]$, and CDC-minus-recursive $-1.2$ $[-4.5, +1.7]$: fixed-size chunking retains the highest recomputation fraction while the content-aware methods remain close. This external analysis checks recomputation cost only and is not pooled with the Wikipedia results: extraction noise is uncharacterised, paragraph boundaries are absent, and lengths were not recalibrated.

\textbf{Retrieval quality (RQ3).}
Under the same encoder, pool, and budget, article-macro-averaged evidence recall for fixed-size, recursive, and CDC chunking is 96.7\%, 96.8\%, and 95.0\% in English, and 97.4\%, 98.3\%, and 98.3\% in Chinese. Mean paired differences range from $-1.9$ to $+0.9$ percentage points; their intervals include or touch zero, with the most negative bound at $-4.5$ points for English CDC minus recursive. Observed mean recall differences are small, but the reported intervals do not rule out a recall decrease; they do not establish equivalence. \rev{These figures, and hence the cost--quality comparison, refer only to the frozen chunk-length targets; the grid measures cost alone. \rev{Fixed-size windows that share 32 or 64 tokens (about 10\% and 20\%) with the previous window recompute within 1.3 points of the non-overlapping windows, embed 10--26\% more input, and in English recall 3.0 and 3.9 points less ($[-7.8, +0.7]$, $[-7.2, -1.1]$), consistent with duplicated text consuming the budget; Chinese recall changes by $-0.9$ and $+0.0$ points.} All three chunkers retrieved every scorable label in 105 of 116 articles. Because they share labels, a label error matters only if it changes whether they agree on it; one label carries 0.4--1.7 points of a language mean, so one or two such errors could shift a difference by as much as observed; we do not rank the chunkers by recall.}

\textbf{Cost in practice.}
$R_c$ represents normalised embedding-input work per revision relative to each chunker's deduplicated full-new-version baseline. On the English cohort, moving from \rev{non-overlapping} fixed-size windows to paragraph-first chunking reduces the mean from 57\% to 38\%; for the 28 small-edit articles, it falls from 46\% to 25\%. These are model-input workload fractions under a document-local previous-version cache and within-version duplicate-input deduplication, not measures of latency or price.

\section{Why Recomputation Happens}
\begin{figure}[t]
  \centering
  \includegraphics[width=\columnwidth]{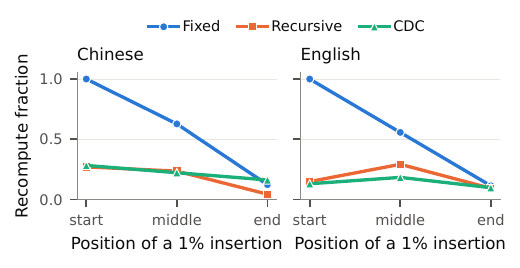}
  \caption{Controlled 1\% insertions at the start, middle, and end of two Chinese and three English development articles: mean recompute fraction by chunker. Cases where any chunker failed are excluded.}
  \label{fig:position}
\end{figure}

\textbf{Position (RQ2).}
Figure~\ref{fig:position} isolates one variable. The same 1\% synthetic insertion is placed at the start, middle, or end of a development article, and each chunker is rerun. In these controlled cases, fixed-size recomputation falls from 100\% at the start to about 12\% at the end: every window after the insertion shifts by the inserted length, so no downstream chunks were reusable. Mean recomputation for paragraph-first recursive chunking and CDC remains below 30\% at all three tested positions, consistent with paragraph boundaries or content-defined cut points resynchronising the partition shortly after the edit. This supports a boundary-drift interpretation of the small-versus-large edit pattern in Figures~\ref{fig:edit-zh} and~\ref{fig:edit-en}: for fixed-size chunking, edit position can affect recomputation beyond the effect of edit size.

\textbf{Verbatim presence.}
On the 17 common-success primary articles in the \rev{24-article development subset (Sec.~\ref{sec:setup})}, we checked whether the complete text of every distinct recomputed embedding input occurs verbatim in the old version. This is the case for 50\% of fixed-size recomputed input tokens in English and 47\% in Chinese, for 33\% and 45\% under CDC, and for 12\% and 37\% under recursive chunking. The complete text of these inputs is present in the old version, but not as a reusable input under the old partition. This is consistent with boundary drift contributing to recomputation. The remaining inputs have no complete verbatim match in the old version; the check does not determine how much of their text would require recomputation under another partition. Because the check does not establish alignment, it supports a boundary-drift interpretation but does not quantify its causal contribution.

\textbf{CDC failures.}
CDC fingerprints whitespace-delimited units, so a long run of Chinese text without whitespace is a single unit; four articles contained units of 403--550 tokenizer units, above the 352 maximum, and the chunker rejected these inputs rather than truncating them. \rev{This is an operational property of CDC: unsegmented Chinese needs word segmentation or a fallback splitter, which would change its boundaries and was not evaluated.} Its cut points must also satisfy minimum and maximum lengths, so after an edit CDC can drift for several chunks before a valid fingerprint boundary is found.

\section{Limitations and Conclusion}\label{sec:limits}
\textbf{Reproducibility.}
Primary results \rev{come from frozen runs} with recorded content hashes; the external analysis is \rev{reproduced by} byte-identical replay. Chunkers never see labels, and the encoder, CDC code, and bootstrap seed are pinned. A research artifact contains scripts, per-article rows, parameter grids, \rev{sensitivity analyses,} benchmark identities, diagnostic articles, and failures; article texts are identified by revision ID and content hash\rev{,} not redistributed.

\textbf{Limitations.}
The cohort comprises 120 adjacent revision pairs from one source, and size bands remain descriptive. Chunk lengths were calibrated separately, but the realised median ratio is 1.17 in both languages and the \rev{order of recursive chunking and CDC changes across grid settings}, so we make no universal claim about recursive versus CDC. The annotations were not independently adjudicated by multiple annotators\rev{, so inter-annotator agreement is unavailable; an author's audit of 60 seeded, stratified second- and third-round labels agreed with 57, and all three disagreements were unanswerable labels whose versions contained relevant information, which removes a label from every chunker's denominator; residual errors could still obscure the few-point recall differences}. The formal token-level edit ratio and amplification factor remain unreported because the natural alignment they require has not been independently audited; we report a proxy and a verbatim-presence diagnostic. Retrieval quality was measured only at \rev{the frozen chunk-length targets; overlap was varied only for fixed-size windows}. A single encoder, retrieval budget, and source were used. We did not obtain a comparable KARA run: in our probes, its token-level chunker did not reproduce the source text byte-for-byte after decoding, a 600-character Chinese paragraph without whitespace exceeded its 64-unit split limit, and its cache scope and cost model differ from the exact single-document payload cache used here. Its absence therefore limits our claims to the three stateless chunkers and is not evidence about KARA's cost or quality.

\textbf{Conclusion.}
On real bilingual revisions, the choice of chunker changes the fraction of a document that must be re-embedded by tens of percentage points, while \rev{at the frozen parameters} observed mean retrieval recall under a fixed budget changes by only a few percentage points, and the cost ranking of the content-aware chunkers varies with chunk-length settings. \rev{Non-overlapping fixed-size} windows have the highest mean recomputation in both languages at the frozen settings and across all tested length-matched grid settings. Practitioners who evaluate chunkers on static retrieval alone should add the recompute fraction, measured under byte-identical reuse and across chunk-length settings, before choosing a chunker for a knowledge base that will be edited.

\section*{Conflict of Interest}
The authors declare that there is no conflict of interest regarding the publication of this paper.

\section*{Acknowledgment}
The authors used ChatGPT and Claude for code development, manuscript revision, and methodological feedback. The authors made the final decisions and take responsibility for this work.

\bibliographystyle{IEEEtran}
\bibliography{refs}

\end{document}